\documentclass[letterpaper, 10 pt, conference]{ieeeconf}  

\IEEEoverridecommandlockouts                              

\usepackage{mathptmx} 
\usepackage{verbatim} 

\usepackage{graphicx}
\usepackage{epstopdf}
\DeclareGraphicsExtensions{.eps}

\usepackage[caption=false,font=normalsize,labelfont=sf,textfont=sf]{subfig}

\usepackage{graphicx}

\usepackage{mathtools,amsthm}

\usepackage{times}
\usepackage{multicol}
\usepackage[bookmarks=true]{hyperref}
\usepackage{amssymb, amsfonts}
\usepackage{cases}

\usepackage{enumerate}

\newcommand{\id}[1][3]{{I}}
\newcommand{\zero}[2]{{0}}
\newcommand{\zeros}[2]{{0}}

\newcommand{\multeqi}[2]{\begin{IEEEeqnarraybox}[][#2]{#1}}
\newcommand{\multeqf}{\end{IEEEeqnarraybox}}

\newcommand{\systemi}[1][rCL]{\left\lbrace\begin{IEEEeqnarraybox}[][c]{#1}}
\newcommand{\systemf}{\end{IEEEeqnarraybox}\right.}

\newcommand{\eqni}[1][rCL]{\begin{IEEEeqnarray}{#1}}
\newcommand{\eqnf}{\end{IEEEeqnarray}}

\newcommand{\nneqni}[1][rCL]{\begin{IEEEeqnarray*}{#1}}
\newcommand{\nneqnf}{\end{IEEEeqnarray*}}

\newcommand{\pmatrixi}{\begin{pmatrix}}
\newcommand{\pmatrixf}{\end{pmatrix}}

\newcommand{\bmatrixi}{\begin{bmatrix}}
\newcommand{\bmatrixf}{\end{bmatrix}}

\newcommand{\smatrixi}{\left[\begin{smallmatrix}}
\newcommand{\smatrixf}{\end{smallmatrix}\right]}

\newcommand{\enumi}{\begin{enumerate}}
\newcommand{\enumf}{\end{enumerate}}

\newcommand{\enumri}{\begin{enumerate}\renewcommand{\theenumi}{\textit{\roman{enumi}}}}
\newcommand{\enumrf}{\end{enumerate}}

\newcommand{\mytheorem}[2]{%
\newtheorem{t#2}{#1}%
\newenvironment{#2}{\begin{t#2}}{\end{t#2}}}

\theoremstyle{plain}
\newenvironment{remark}[1][Remark]{\begin{trivlist}
\item[\hskip \labelsep {\bfseries #1}]}{\end{trivlist}}

\mytheorem{Theorem}{theorem}
\mytheorem{Lemma}{lemma}
\mytheorem{Proposition}{proposition}
\mytheorem{Corollary}{corollary}
\mytheorem{Assumption}{assumption}
\mytheorem{Definition}{definition}
\mytheorem{Property}{property}

\usepackage{paralist}
\usepackage{color}

\title{\LARGE \bf
On-line Joint Limit Avoidance for Torque Controlled Robots \\ by Joint Space Parametrization
}

\author{Marie Charbonneau, Francesco Nori, and Daniele Pucci$^{1}$
\thanks{$^{1}$ Marie Charbonneau, Francesco Nori and Daniele Pucci are with the iCub Facility Department, Istituto Italiano di Tecnologia, Via Morego 30, 16163, Genova, Italy 
        {\tt\small (email: name.surname@iit.it)}
        }
}

\begin{document}

\maketitle
\thispagestyle{empty}
\pagestyle{empty}


\begin{abstract}

This paper proposes control laws ensuring the stabilization of a time-varying desired joint trajectory and joint limit avoidance in the case of fully-actuated manipulators. The key idea is to perform a parametrization of the feasible joint space in terms of exogenous states. It follows that the control of these states allows for joint limit avoidance. One of the main outcomes of this paper is that position terms in control laws are replaced by parametrized terms. Stability and convergence of time-varying reference trajectories obtained with the proposed method are demonstrated to be in the sense of Lyapunov. The introduced control laws are verified by carrying out experiments on two degrees-of-freedom of the torque-controlled humanoid robot iCub.

\end{abstract}


\section{INTRODUCTION}


Nonlinear control of unconstrained fully-actuated  manipulators is no longer a theoretically challenging problem for the control community. Position, velocity, and torque-control based algorithms have long been analysed with back-stepping and feedback linearization tools, and have proved to be effective in numerous applications, e.g.~\cite{Isenberg2010, Luo2013, Mason2014}. The control problem associated with robotic manipulators, however, rapidly becomes challenging when motion and actuation constraints must be satisfied. This paper contributes along this line by proposing control solutions to ensure joint limit avoidance for torque-controlled manipulators.


The problem of ensuring joint limit avoidance is not new to the robotics community.
For instance, a variety of  methods were developed 
in path planning, such as weighted least norm solutions~\cite{Chan1995}, damped least square solution of inverse kinematics~\cite{Na2008}, Lyapunov-based methods~\cite{Chen2006}, neural networks~\cite{Zhang2003} and~\cite{Assal2005}, or using a time-varying weight matrix in the inverse kinematics~\cite{Kermorgant2011}. Nonetheless, generating reference trajectories that satisfy the physical limits does not imply that the joint positions will evolve within these limits.

In the case of redundant manipulators, on-line joint limit avoidance may be attempted by using the stack-of-tasks approach. In fact, the control objective associated with redundant manipulators is usually the stabilisation of the robot end-effector, and the solutions associated with this task may not be unique. One can exploit redundancy by defining a secondary low-priority task, in charge of keeping the joints away from limits and acting onto the null space of the main task~\cite{Fukumoto2004,Jamone2013,Fiacco2013}. One of the main drawbacks of this approach is that there is no theoretical guarantee that the joint evolutions always belong to the feasible domain. Also, the two-layer prioritization may lead to undesired robot behaviour, due to the projection onto the null space of the control action in charge of ensuring joint limit avoidance. 

Among the most widely used methods for joint limit avoidance of redundant manipulators is the gradient projection based technique~\cite{Liegeois1977,Chen2002,Masahide2010,Marey2010}. This approach defines a criterion, such as a function maximizing the distance between joint positions and their limits. The gradient of this function is then projected onto the null space projection matrix of the Jacobian, allowing to move the joints away from limits without affecting the end-effector position. This method has a few drawbacks, as it does not guarantee minimization of the criterion for each individual joint, and additional coefficients need to be used to properly tune the self-motion magnitude.

In humanoid whole-body motion control, unilateral virtual springs and spring-dampers have been implemented around joint limits to generate torques repelling from the bounds~\cite{Dietrich2011,Moro2013}. Another possibility for humanoid robots is to solve whole-body motion as an optimization problem with inequality constraints corresponding to joint limits~\cite{Herzog2013,Tassa2014,Feng2014,Hopkins2015}. In all of these works, however, the theoretical guarantee of the stability and convergence properties associated with the evolution of the system is still missing.

This paper presents a novel nonlinear control algorithm that ensures joint limit avoidance for fully-torque-controlled manipulators. The control objective is the asymptotic stabilization of a desired joint reference trajectory. It is achieved by ensuring that the evolution of the joints always remains within the associated physical bounds. The essence of the proposed control algorithm is to parametrize the feasible joint space in terms of exogenous states, and then the control of these states allows for the achievement of joint limit avoidance. Stability and convergence, when the desired joint trajectory is feasible, are shown by means of an analysis based on Lyapunov theory. One of the main outcomes of this paper is to define nonlinear position feedback terms which can be used in lieu of classical position correction terms when a desired joint trajectory must be followed. The control laws proposed here are reminiscent of those obtained by applying \emph{barrier-function} based control approaches~\cite{Ngo2006, Prajna2004, Tee2009, Wieland2007}, but they are derived from a different perspective. The presented control laws are verified by controlling two degrees-of-freedom on the torque-controlled iCub humanoid robot~\cite{Metta2008}. 

The paper is organised as follows. Section~\ref{sec:background} introduces the notation, standard models and controllers for fixed-base manipulators. Section~\ref{sec:mainResults} presents the main idea for achieving joint limit avoidance. Section~\ref{sec_jointSpaceControl} presents and discusses the control laws for stabilizing a desired joint trajectory and for ensuring joint limit avoidance. Section~\ref{sec:simulationsAndExperiments} describes the experiments carried out to validate the approach. The paper is concluded by remarks and perspectives.


\section{BACKGROUND} \label{sec:background}

\subsection{Notation}

The following notation is used throughout the paper.
\begin{itemize}
	\item The set of real numbers is denoted by $\mathbb{R}$.
	\item The Euclidean norm of a vector of coordinates $v \in \mathbb{R}^n$ is denoted by $|v|$.
	\item Given a time function $f(t) \in \mathbb{R}^n$, its first- and second-order time derivatives are denoted by $\dot{f}(t)$ and $\ddot{f}(t)$, respectively.
\end{itemize}

\subsection{System Modelling}

The Lagrangian derivation of the equations of motion of a robotic manipulator with $n$ degrees of freedom yields a model of the following form~\cite{Siciliano2008}:

\begin{equation}
    \label{eq_SystemModel}
	M(q) \ddot{q} + C(q,\dot{q})\dot{q} + G(q) = \tau 
\end{equation}
where $q \in \mathbb{R}^n$ is the vector of generalized coordinates of the mechanical system, 
$M(q) \in \mathbb{R}^{n \times n}$, $C(q,\dot{q}) \in \mathbb{R}^{n \times n}$ and $G(q) \in \mathbb{R}^n $ are the inertia matrix, Coriolis matrix and gravity torques, respectively, and $\tau$ is the vector of input torques.

The following properties of model (\ref{eq_SystemModel}) are assumed~\cite{Siciliano2008}:
\begin{property}
    \label{hp:mPositiveDefinite}
    The inertia matrix $M$ is bounded and symmetric positive definite for any $q$.
\end{property}
\begin{property}
    \label{hp:mM2CSkew}
     The matrix $\dot{M} - 2C$ is skew-symmetric.
\end{property}

\subsection{Control objective and recalls on classical control schemes}

Let $q_d(t) \in \mathbb{R}^n$ denote a twice differentiable time function representing the desired trajectory for the joint configuration~$q$. Throughout the paper, we assume that: 
 \begin{assumption}
    \label{hp:reference}
    The reference trajectory $q_d(t)$ is such that its first and second order time derivatives are well-defined and bounded $\forall t \in \mathbb{R}^+$.
 \end{assumption}

Then, the control objective is defined as the asymptotic stabilization of the tracking error
\begin{IEEEeqnarray}{RCL}
    \label{eq_trackingError}
    \tilde{q} &=& q - q_d
\end{IEEEeqnarray}
to zero. To achieve this objective, classical control laws can be applied. For instance, passivity-based controllers are known to work robustly against modelling and actuation errors \cite[ch. 8.5.1 p. 328]{Siciliano}, and the associated  law writes
\begin{IEEEeqnarray}{RCL}
    \label{eq_GravityCompensation}
    \tau &=&  M(q) \ddot{q}_d + C(q, \dot{q}) \dot{q}_d +G(q) - K_P \tilde{q} - K_D \dot{\tilde{q}} 
\end{IEEEeqnarray}
with $K_P$ and $K_D$ two symmetric, positive definite matrices representing  proportional and derivative control gains.
Applying control law (\ref{eq_GravityCompensation}) to system (\ref{eq_SystemModel}) results in bounded trajectories of the closed-loop dynamics and convergence of the tracking error to zero, for any initial condition $(q, \dot{q})(0)$.

However, overshoots of the joint evolutions, or external forces applied to the system, may cause the robot to hit joint limits. What follows  proposes a solution to the problem of joint limit avoidance while retaining stability and convergence of the tracking error $\tilde{q}$ to zero.

\section{JOINT SPACE PARAMETRIZATION} \label{sec:mainResults}


Let $q_{min}, q_{max} \in \mathbb{R}^n$ denote the vectors defining the minimum and maximum values of the joint coordinates $q$. We define the feasible space $\mathbb{Q}$ for the joint coordinates as:
\begin{IEEEeqnarray}{RCL}
\label{eq_feasibleDomain}
    \mathbb{Q} &:=& \{q\in \mathbb{R}^n : 
    q_{{min}_i} < q_i < q_{{max}_i} 
    \ \forall i = 1, \cdots, n \}.
\end{IEEEeqnarray}
The control objective is then the global asymptotic stabilization of the tracking error~\eqref{eq_trackingError} to zero while ensuring that $q(t) \in \mathbb{Q} \quad \forall t \geq 0$.

To ensure that the variable $q$ always belongs to $\mathbb{Q}$, one may parametrize the feasible configuration space. Let $\xi~\in~\mathbb{R}^n$ denote an exogenous variable. Then, we propose here to consider the following parametrization of the space $\mathbb{Q}$:
\begin{IEEEeqnarray}{RCL}
    \label{eq_parametrization}
    q(\xi) &:=& \delta \tanh(\xi) + q_0  
\end{IEEEeqnarray}
with $q_0 := \frac{q_{max} + q_{min}}{2}$,
$\delta := \text{diag}\left(\frac{q_{max} - q_{min}}{2} \right)$.

$\text{diag}(\cdot):~\mathbb{R}^n~\rightarrow~\mathbb{R}^{n\times n}$ is the operator that given a vector $x\in\mathbb{R}^n$ returns a diagonal matrix having on the diagonal the elements of the vector $x$, and $\tanh(\xi): \mathbb{R}^n \rightarrow  \mathbb{R}^n$. As a consequence of the hyperbolic function nature, one clearly has that $q(\xi)~\in~\mathbb{Q}~\quad\forall~\xi~\in~\mathbb{R}^n$. We now make the following assumption.

 \begin{assumption}
    \label{hp:desTrajAndLimits}
    Each joint coordinate $q_i$ possesses a \emph{free motion domain} different from zero, i.e.
    $q_{{max}_i}~-~q_{{min}_i}~>~0 \quad \forall~i~=~1,~\cdots,~n$
    and the reference trajectory $q_d(t)$ is feasible, 
    i.e. $q_d(t) \in \mathbb{Q} \quad t \geq 0$.
 \end{assumption}

As a consequence of the above assumption, one can evaluate the desired trajectory $\xi_d(t)$ for the variable $\xi$ via Eq.~\eqref{eq_parametrization}, i.e.
$\xi_d(t) := \tanh^{-1} \left[ \delta^{-1} (q_d(t)- q_0) \right]$,
and define the tracking error as $\tilde{\xi} := \xi - \xi_d$.

The main idea presented in this paper is to conceive feedback control laws for the asymptotic stabilization of $\tilde{\xi}$ to zero, which, relying on the nature of the parametrization~\eqref{eq_parametrization}, would imply that $q(t) \in \mathbb{Q} \ \forall t \geq 0$. 

Now, it is observed that the relationship~\eqref{eq_parametrization} can be viewed as a change of variable $\xi \rightarrow q$. So, the  equations of motion~\eqref{eq_SystemModel} can be written in terms of $\xi$. To this purpose, note that
\begin{IEEEeqnarray}{RCL}
    \dot{q} &=& J(\xi) \dot{\xi} \IEEEyessubnumber
    \label{eq_qDot} \\
    \ddot{q} &=& J(\xi) \ddot{\xi} + \dot{J}(\xi, \dot{\xi}) \dot{\xi} \IEEEyessubnumber
    \label{eq_qDDot}
\end{IEEEeqnarray}
with $J \in \mathbb{R}^n$ a diagonal matrix whose $i-$th element is given by 
\begin{IEEEeqnarray}{RCL}
    \label{eq_jacobian_i}
     J_i(\xi) = \delta_i (1 - \tanh^2 (\xi_i)),
\end{IEEEeqnarray}
and $\delta_i = (q_{{max}_i} - q_{{min}_i})/2$.
It is important to observe that if Assumption~\ref{hp:desTrajAndLimits} holds, which implies that $\delta_i \ne 0$ $\forall i$, then
\begin{IEEEeqnarray}{RCL}
    \label{eq_detJacobian}
     \det(J(\xi)) \ne 0 \quad \forall \xi \in \mathbb{R}^n.
\end{IEEEeqnarray}
Then, as long as the joint configurations belong to  $\mathbb{Q}$, the equations of motion~\eqref{eq_SystemModel} can be written as
\begin{IEEEeqnarray}{LCL}
    \label{eq_SystemModelXi}
	M_\xi(\xi) \ddot{\xi} + C_\xi(\xi,\dot{\xi})\dot{\xi} + G_\xi(\xi) &=& \tau_\xi 
\end{IEEEeqnarray}
with 
\begin{IEEEeqnarray}{RCL}
    \label{componentsNewDyanmcis}
    M_\xi(\xi) &=& J^T(\xi) M J(\xi) \IEEEyessubnumber \\
    C_\xi(\xi,\dot{\xi}) &=& J^T(\xi) (M \dot{J}(\xi,\dot{\xi}) + CJ(\xi)) \IEEEyessubnumber \\
    G_\xi(\xi) &=& J^T(\xi) G \IEEEyessubnumber \\
    \tau_\xi  &=& J^T(\xi) \tau \IEEEyessubnumber
    \label{eq_tauXi}
\end{IEEEeqnarray}
Observe that the matrix $J(\xi)$ is bounded for any $\xi$. Then, it is straightforward to verify the following two properties of model~\eqref{eq_SystemModelXi}, which reflect properties~\ref{hp:mPositiveDefinite} and~\ref{hp:mM2CSkew} of model~\eqref{eq_SystemModel}.
\begin{property}
    \label{hp:mxiPositiveDefinite}
    The inertia matrix $M_\xi$ is bounded and symmetric positive definite for any $\xi$.
\end{property}
\begin{property}
    \label{hp:mM2CxiSkew}
     The matrix $\dot{M}_\xi - 2C_\xi$ is skew-symmetric.
\end{property}

\section{JOINT SPACE CONTROL\\ WITH JOINT LIMIT AVOIDANCE}
\label{sec_jointSpaceControl}
In this section, we present and discuss control laws for stabilizing a desired joint trajectory $q_d(t)\in \mathbb{Q} \ \forall t$ that ensure joint limit avoidance. 

Let us first remark an important fact. Once the system dynamics~\eqref{eq_SystemModel} is transformed  into the form~\eqref{eq_SystemModelXi}, any controller ensuring that the variable $\xi$ is bounded would also imply that the joint trajectories belong to the feasible joint space $\mathbb{Q}$. For instance, the computed-torque-like control strategy can be applied assuming $\tau_\xi$ as control input, and this would ensure that $\xi$ is bounded and, in turn, that $q(t) \in \mathbb{Q} \ \forall t$. 

Extending the passivity-based control strategy~\eqref{eq_GravityCompensation} to system~\eqref{eq_SystemModelXi} requires some close attention. The major technical difficulties reside in the fact that the variable change $\xi \rightarrow q$ is not one-to-one for any $q \in \mathbb{R}^n$, in the sense that if $q$ is outside the feasible joint space, then $\nexists$ $\xi$ such that $q = q(\xi)$. This implies that the matrix $M_\xi$ tends to zero when joint trajectories approach their limits. The extension, however, is presented in the next lemma.
\begin{lemma}
Assume that Property~\ref{hp:mPositiveDefinite} and Assumption~\ref{hp:reference} hold. 
     Apply to system~\eqref{eq_SystemModel} the following control law:
     \begin{IEEEeqnarray}{RCL}
        \label{eq_controlTauXi}
         \tau_{\xi} = M_{\xi} \ddot{\xi}_d + C_{\xi}(\xi, \dot{\xi}) \dot{\xi}_d +G_{\xi}(\xi) - K_P \tilde{\xi} - K_D \dot{\tilde{\xi}}.
     \end{IEEEeqnarray}
     Then, the following results hold.
    \begin{enumerate}
        \item The equilibrium point $\left(\tilde{\xi},\dot{\tilde{\xi}}\right) = (0,0)$ of the closed loop dynamics~\eqref{componentsNewDyanmcis}-\eqref{eq_controlTauXi} is globally asymptotically stable;
        \item If $q(0) \in \mathbb{Q}$, then $q(t) \in \mathbb{Q} \ \forall t \geq 0$.
    \end{enumerate}
\end{lemma}

Proof is given in Appendix. The control law~\eqref{eq_controlTauXi} ensures that the joint evolutions $q(t)$ belong to the feasible space $\mathbb{Q}$ for any time $t$, provided that the initial condition $q(0)$ belongs to this space. 

The proof of this law exploits the passivity of the system dynamics expressed by Properties~\ref{hp:mxiPositiveDefinite} and~\ref{hp:mM2CxiSkew}, and it must deal with the additional technicality that the mass matrix $M_\xi$ tends to zero when the joint evolutions get closer to the joint limits.
Observe the similarity between the control laws~\eqref{eq_controlTauXi} and~\eqref{eq_GravityCompensation}. All these similarities constitute the interest of the proposed parametrization~\eqref{eq_parametrization}.

The control torques $\tau$ can be directly evaluated from~\eqref{eq_controlTauXi} and~\eqref{componentsNewDyanmcis}, that is 
\begin{IEEEeqnarray}{RCL}
    \label{eq_torqueBounded}
     \tau &=& MJ(\xi) \ddot{\xi}_d + 
     \left(M \dot{J}(\xi,\dot{\xi}) + CJ(\xi) \right)\dot{\xi}_d +G \nonumber \\ 
     &-& J^{-1}(\xi) K_P \tilde{\xi} - J^{-1}(\xi) K_D \dot{\tilde{\xi}}.
\end{IEEEeqnarray}
Therefore, note that the similarities between the control laws ~\eqref{eq_torqueBounded} and~\eqref{eq_GravityCompensation} increase when the reference trajectory is a set point, i.e. $\dot{\xi}_d = \ddot{\xi}_d = 0$, which implies that \begin{IEEEeqnarray}{RCL}
     \tau &=& G
     - J^{-1}(\xi) K_P \tilde{\xi} - J^{-1}(\xi) K_DJ^{-1}(\xi) \dot{\xi}. \nonumber
\end{IEEEeqnarray}
$J(\xi)$ being positive definite, one can choose the control gains $K_P = J(\xi) K'_P$, $K_D = J(\xi) K'_DJ(\xi)$ and $K'_P, K'_D > 0$ without destroying stability and convergence.

Then, in the case of set points, the main difference between classical control algorithms and the proposed control solutions reside in the feedback position terms:
\begin{IEEEeqnarray}{RCL}
    \label{eq_torqueBoundedSetPoint}
     \tau &=& G(q) 
     -  K'_P \tilde{\xi} - K'_D \dot{q},
\end{IEEEeqnarray}
although theoretical guarantee of the stability and convergence of the control law~\eqref{eq_torqueBoundedSetPoint} is missing at this point.

Eq.~\eqref{eq_torqueBoundedSetPoint} suggests that given the classical control scheme~\eqref{eq_GravityCompensation}, joint limit avoidance can be attempted by substituting the feedback correction term $-K_p\tilde{q}$
with either $- J^{-1}(\xi) K_P \tilde{\xi}$
or $-K_p\tilde{\xi}$,
since the associated control laws can be shown to ensure joint limit avoidance. This is a general procedure that may be attempted any time joint limits must be taken into account, and the control laws contain feedback position terms. 

\begin{remark}
The implementation on a real platform of the control law~\eqref{eq_torqueBoundedSetPoint} 
requires close attention since they involve singularities of the variable $\xi$. These singularities may cause high-value for the torque input. Then, we suggest to use properly defined saturation functions to avoid explosions of the variable $\xi$ depending on the torque limits of the underlying platform. Simulations and experiments we carried out, however, tend to show that the feedback correction terms
$-K_p\tilde{\xi}$
do not cause sharp, disruptive variations of the control variable $\tau$, and we thus suggest the use of $-K_p\tilde{\xi}$
over the other presented control laws.
\end{remark}

\section{EXPERIMENTAL RESULTS} \label{sec:simulationsAndExperiments}

The proposed control law was tested in simulation and in experiments with the humanoid robot iCub~\cite{Metta2008}. Simulations have been performed with a 2 degrees-of-freedom manipulator, verifying the convergence and stability properties of the approach. However, they are omitted here due to space constraints. Experiments with the iCub, on the other hand, allowed to observe compliance and robustness obtained with the proposed controller.

\begin{figure}[!t]
    \centering
    \includegraphics[width=2.2in, angle =0]{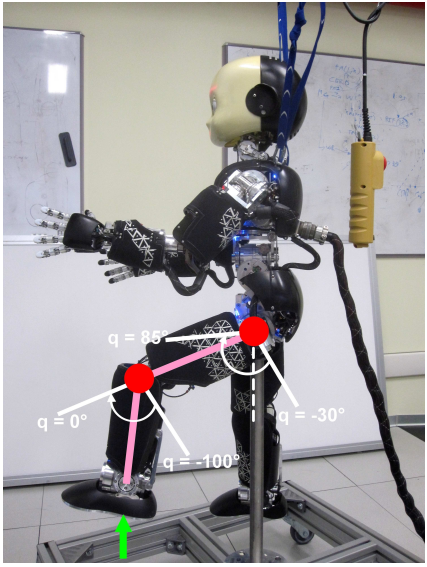}
    \caption{iCub leg setup used for the experiments. The red circles identify the hip and knee joints, while the white marks indicate joint limits. The green arrow shows the external force applied in Experiment 3.}
    \label{fig:iCub}
\end{figure}

The leg of the iCub was used, forming a 2 degrees-of-freedom manipulator with rotational joints at the hip and knee (see Fig.~\ref{fig:iCub}). The ankle joint was kept fixed with a position controller. Joints were bounded within limits set to $[-30, 85] deg$ for the hip and $[-100, 0] deg$ for the knee. Moreover, the leg is equipped with a 6 axis force-torque sensor in the foot, as well as position and torque sensors in each joint. Joint torques obtained from the control laws are stabilized by a low-level joint torque controller. As discussed in the remark above, to avoid singularity issues, saturation was defined for the variable $\xi$ at a value of $100$.

Note that a small approximation in the control laws was made, due to limitations of the software associated to the iCub: it allows for the evaluation of bias forces $G(q)$ and $C(q, \dot{q}) \dot{q}$ acting on each joint. However, it does not allow for the computation of $C(q, \dot{q})$ itself. 
As a solution, $C(q, \dot{q}_d) \dot{q}_d$ and $C(q, J \dot{\xi}_d) J \dot{\xi}_d$ were used in~\eqref{eq_GravityCompensation} and~\eqref{eq_torqueBounded}. The impact of this approximation is minor, since joint velocities used in the experiments are small and $C$ is kept to a low value. 

\subsection{Experiment 1 - constant reference position}

The first experiment consisted in reaching a constant joint reference position $q_d = [-18.5, -10.0] deg$ for the hip and knee joints respectively, from a given initial position $q(0)~=~[-14, -60] deg$ and initial velocity $\dot{q}(0) = [0, 0]$. The proportional and derivative gain matrices $K_p$ and $K_d$ were chosen as diagonal matrices with stiffness values of $[20, 10] N/m$ associated to hip and knee joints respectively, and damping values of $[0, 0]$.

\begin{figure}[!t]
    \centering
    \subfloat{\includegraphics[width=1.6in, trim={0 0.1cm 0 0.1cm},clip]{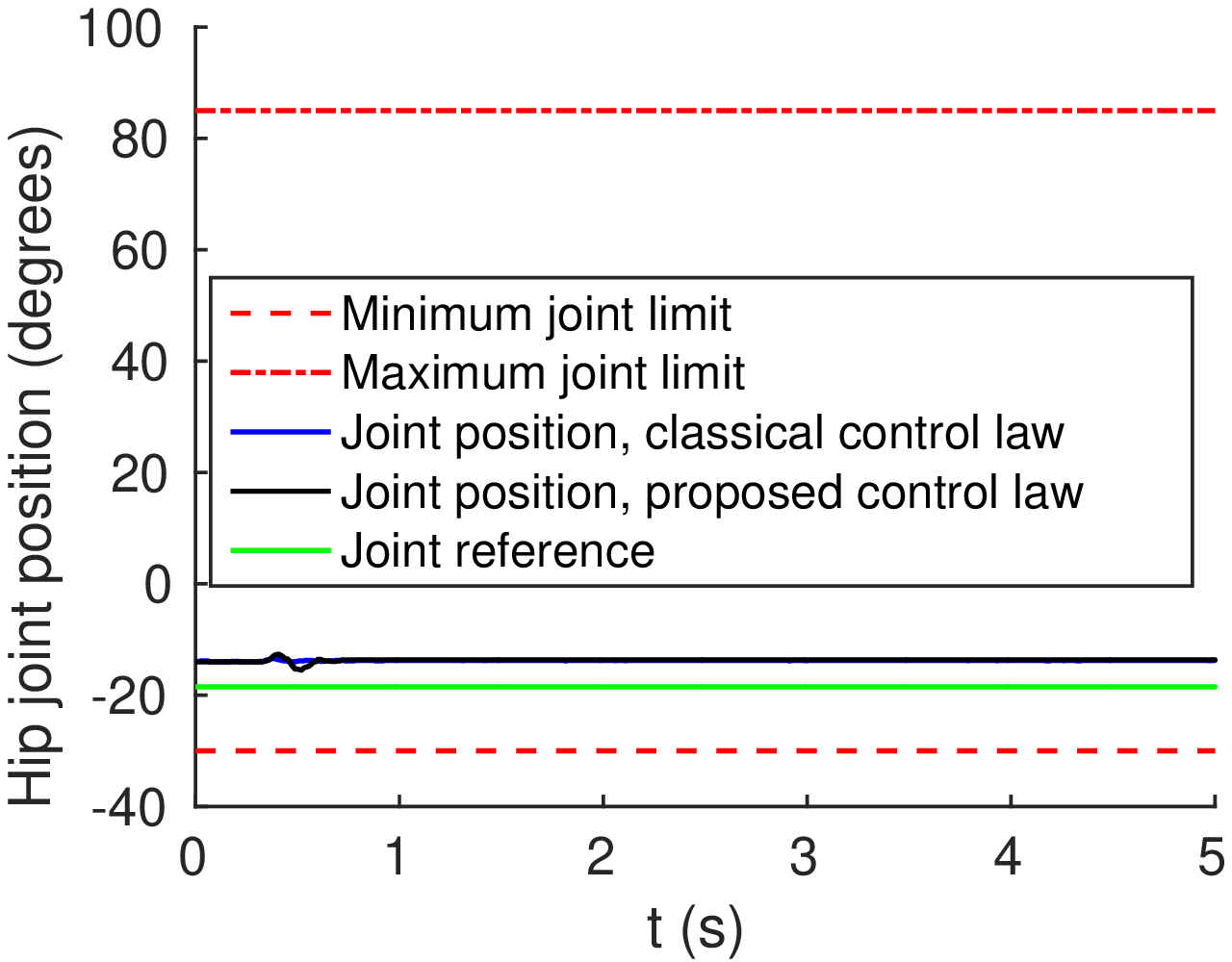}
    \label{fig_hipSetPointPositions}}
    ~
    \subfloat{\includegraphics[width=1.6in, trim={0 0.1cm 0 0.1cm},clip]{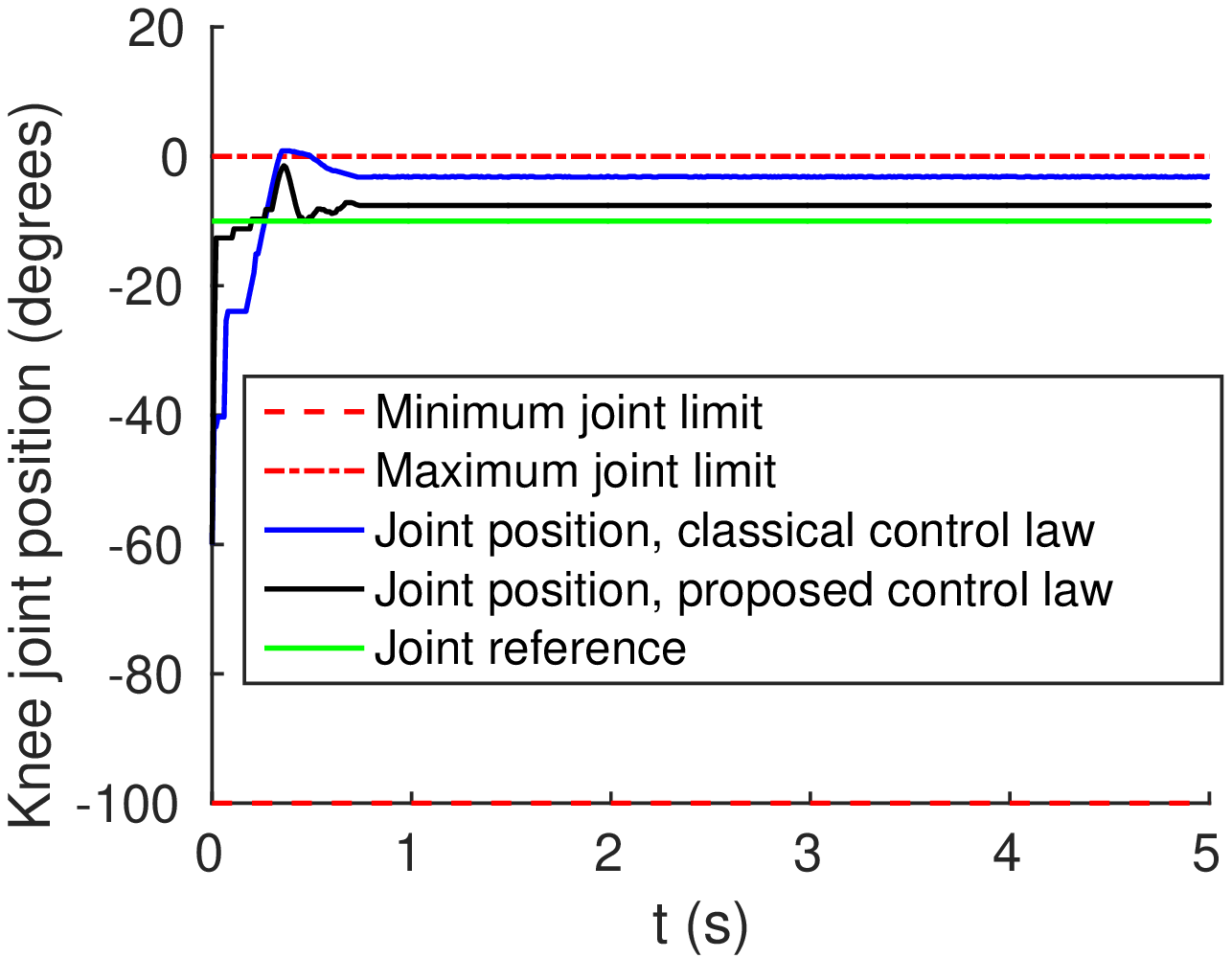}
    \label{fig_kneeSetPointPositions}}
    \hfil 
    \subfloat{\includegraphics[width=1.6in, trim={0 0.1cm 0 0.6cm},clip]{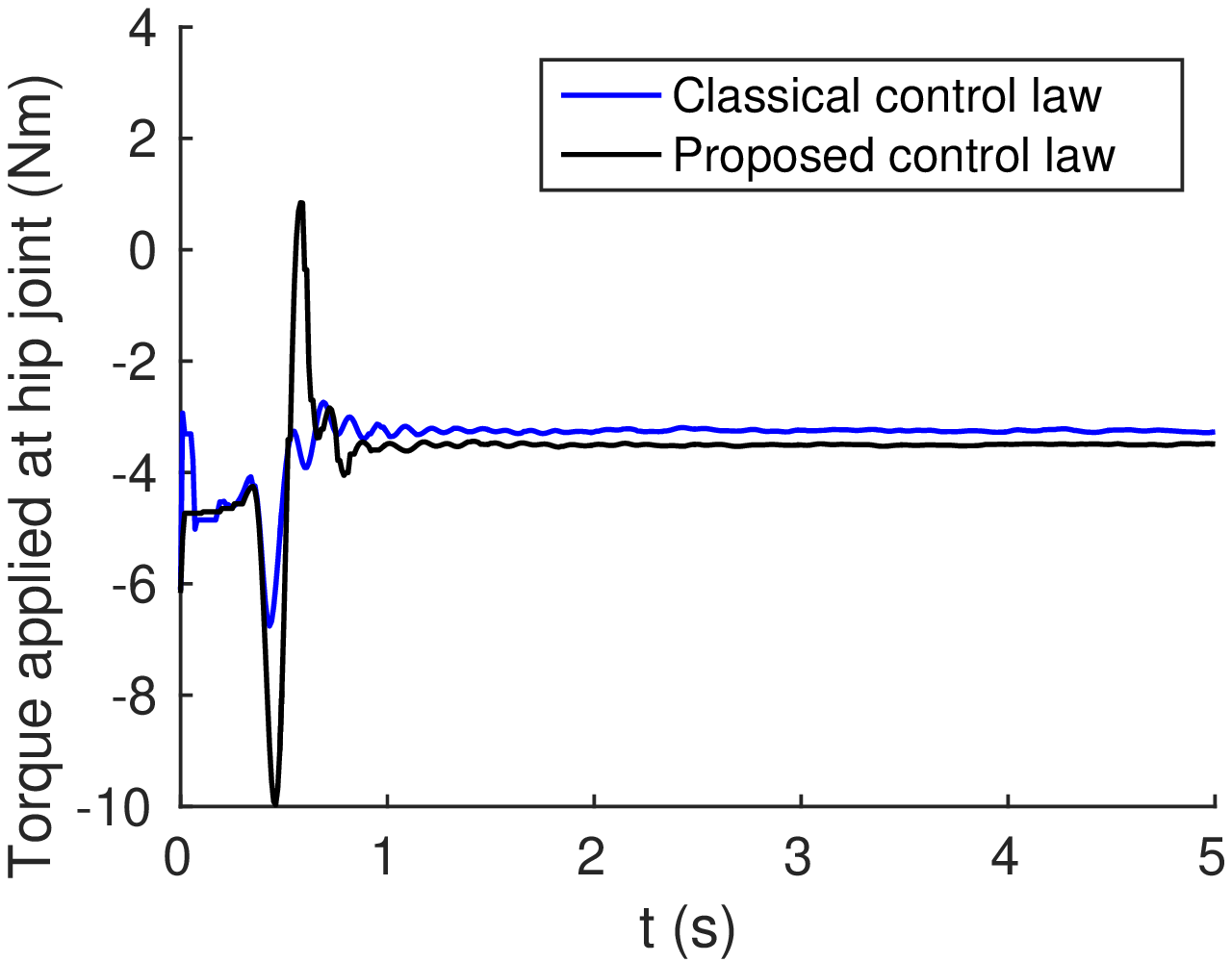}
    \label{fig_hipSetPointTorques}}
    ~
    \subfloat{\includegraphics[width=1.6in, trim={0 0.1cm 0 0.6cm},clip]{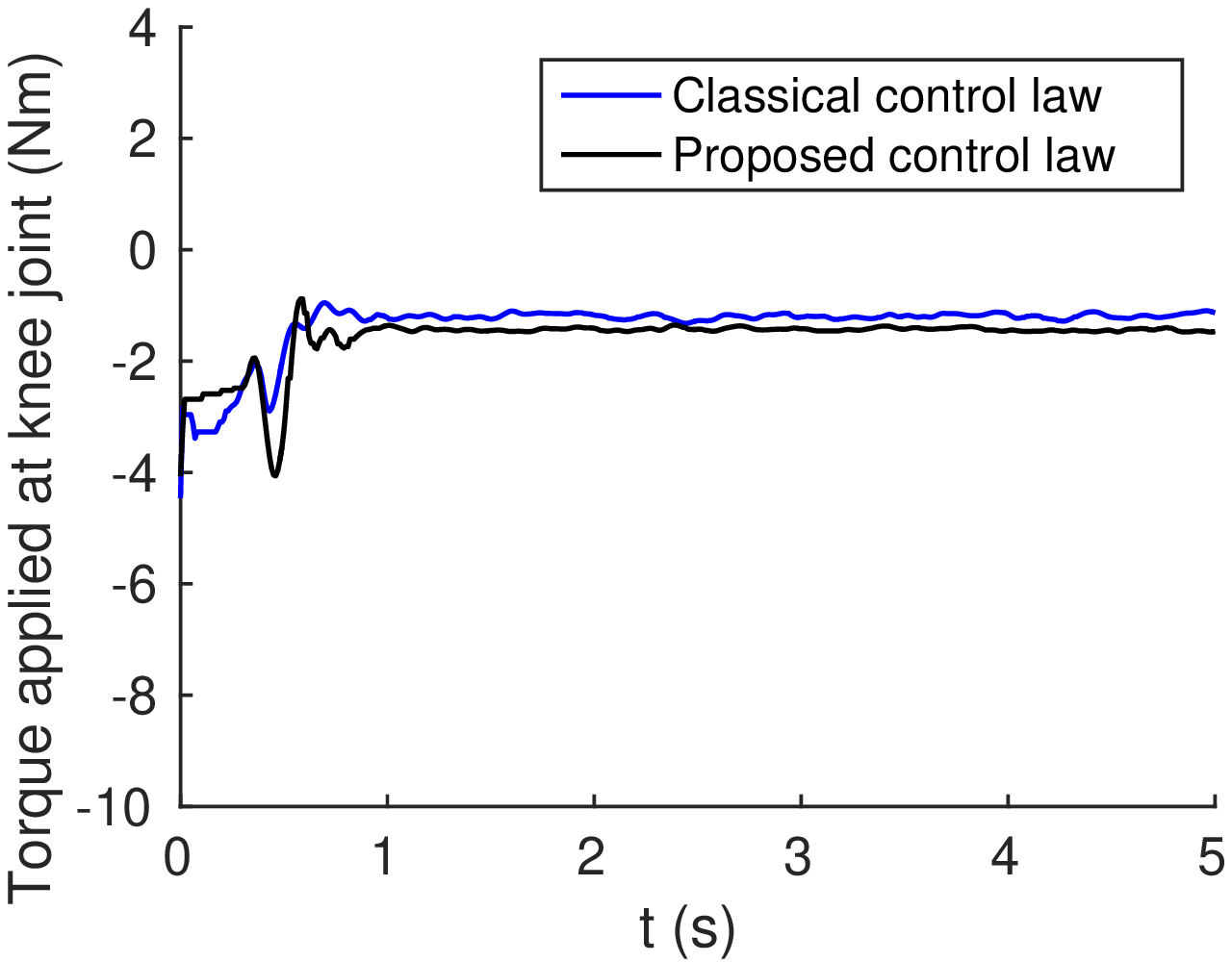}
    \label{fig_kneeSetPointTorques}}
    \caption{Hip and knee joint trajectories and torque, resulting from Experiment 1.}
    \label{fig:ExperimentSetPoint}
\end{figure}

Fig.~\ref{fig:ExperimentSetPoint} shows the evolution of the joint positions and control torques obtained in the experiment. Overshoot causes the knee joint to overpass its limit when using the classical passivity-based law~\eqref{eq_GravityCompensation}, while the knee joint remains within limits when using the proposed control law~\eqref{eq_torqueBounded}.

Note that the iCub platform is equipped with a low-level torque control loop in charge of stabilizing any desired joint~\cite{Fumagalli2010, Fumagalli2012}; it compensates for friction effects, but with some imperfections, and some viscous friction remains present. The fact that the tracking error does not converge to zero is thus mainly due to imperfect tracking of this low-level loop and to unmodeled friction effects.

\subsection{Experiment 2 - sinusoidal reference position}

A second experiment consisted in tracking time-varying sinusoidal joint reference positions of the form
    $q_d (t)~=~\frac{\delta}{r} \sin (\omega t + \rho) + q_0$.
    
Parameters were set to $r = 1.1$, and for the hip and knee joints respectively: $\omega = [0.25, 0.65]$ and $\rho = [0; -\pi/2]$. Initial conditions were $q(0) = [-14, -60] deg$, $\dot{q}(0) = [0, 0]$. The proportional and derivative gain matrices $K_p$ and $K_d$ were chosen as diagonal matrices with stiffness values of $[68, 17] N/m$ associated to hip and knee joints respectively, and damping values of $[0, 0]$.

\begin{figure}[!b]
    \centering
    \subfloat{\includegraphics[width=1.6in, trim={0 0.1cm 0 0.1cm},clip]{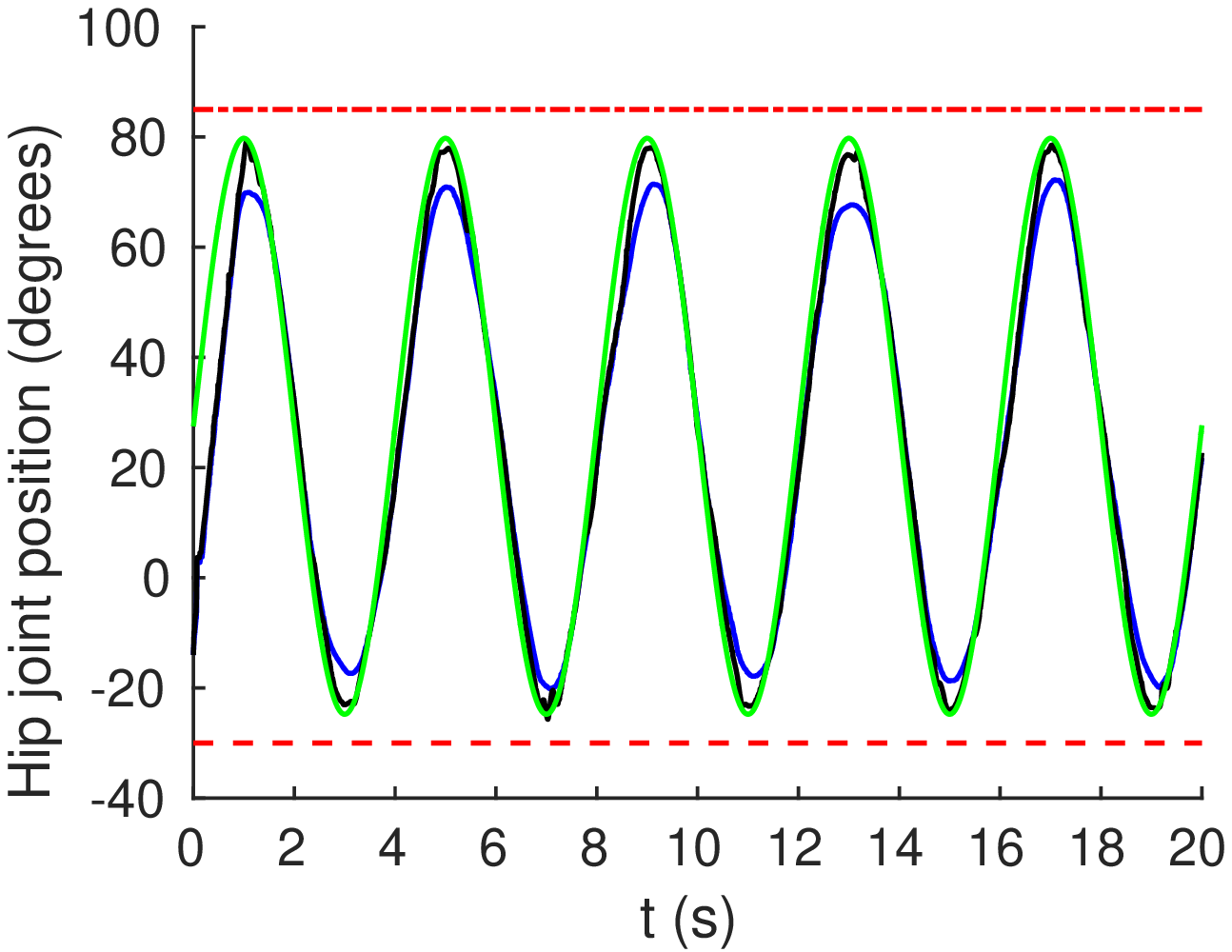}
    \label{fig_hipSinusPositions}}
    ~
    \subfloat{\includegraphics[width=1.6in, trim={0 0.1cm 0 0.1cm},clip]{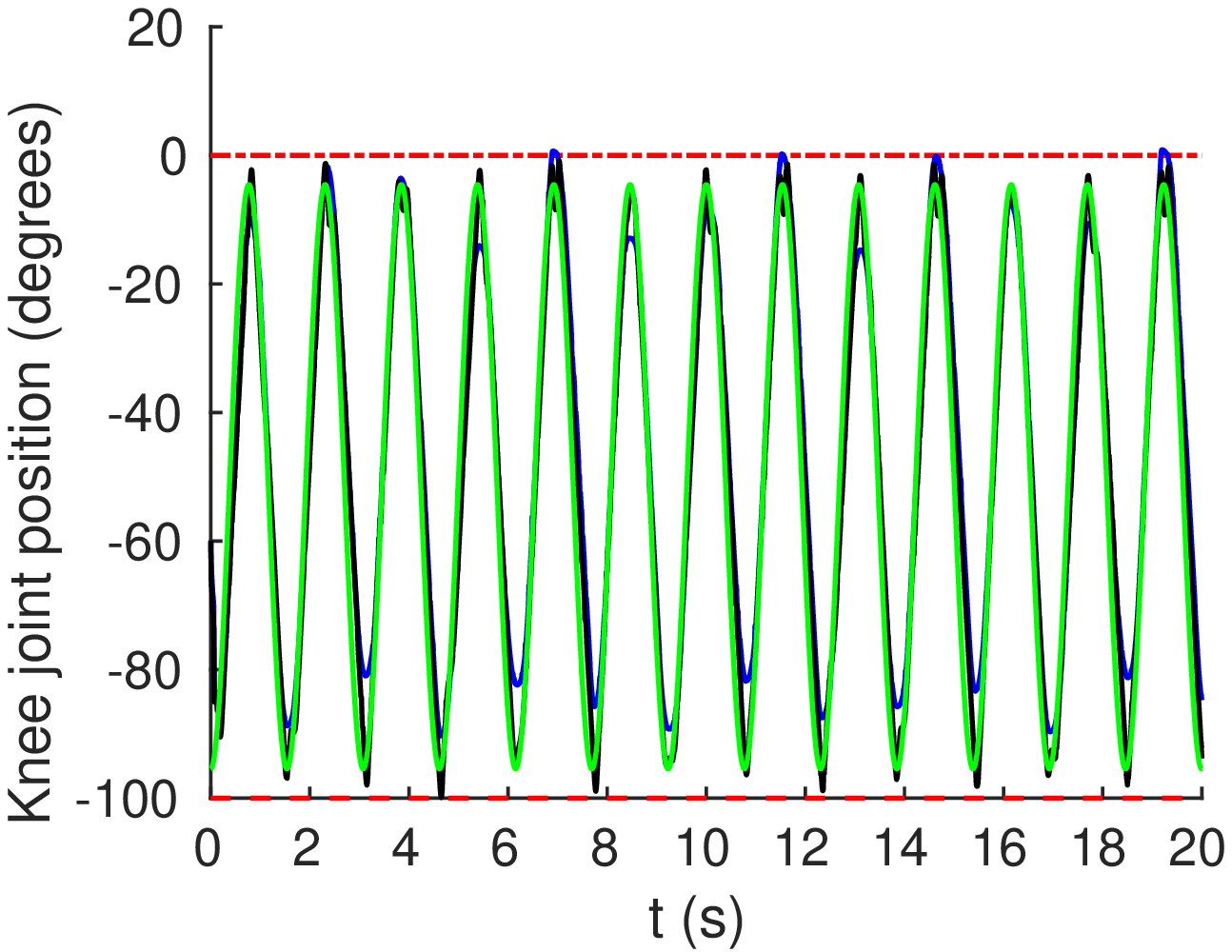}
    \label{fig_kneeSinusPositions}}
    \hfil 
    \subfloat{\includegraphics[width=1.6in, trim={0 0.1cm 0 0.6cm},clip]{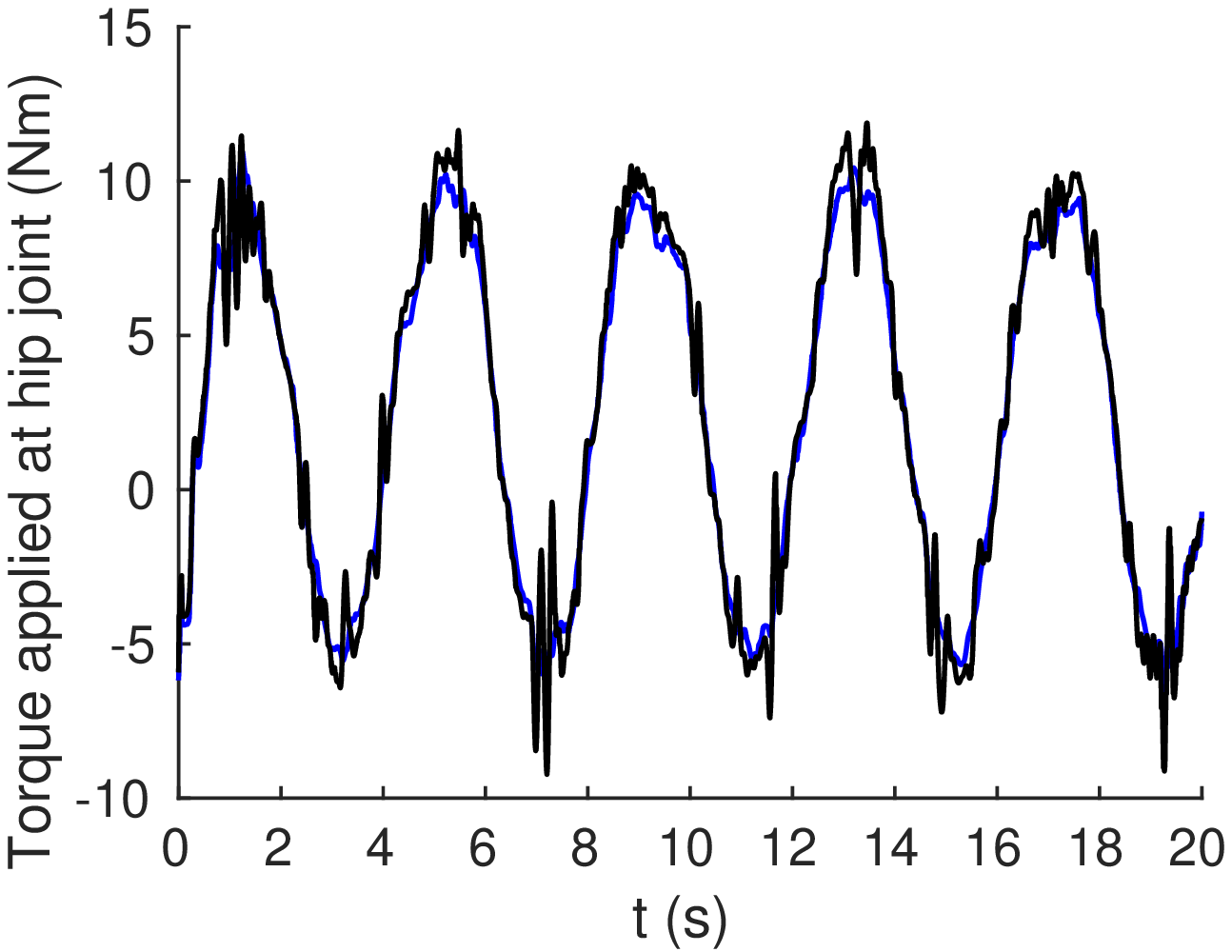}
    \label{fig_hipSinusTorques}}
    ~
    \subfloat{\includegraphics[width=1.6in, trim={0 0.1cm 0 0.6cm},clip]{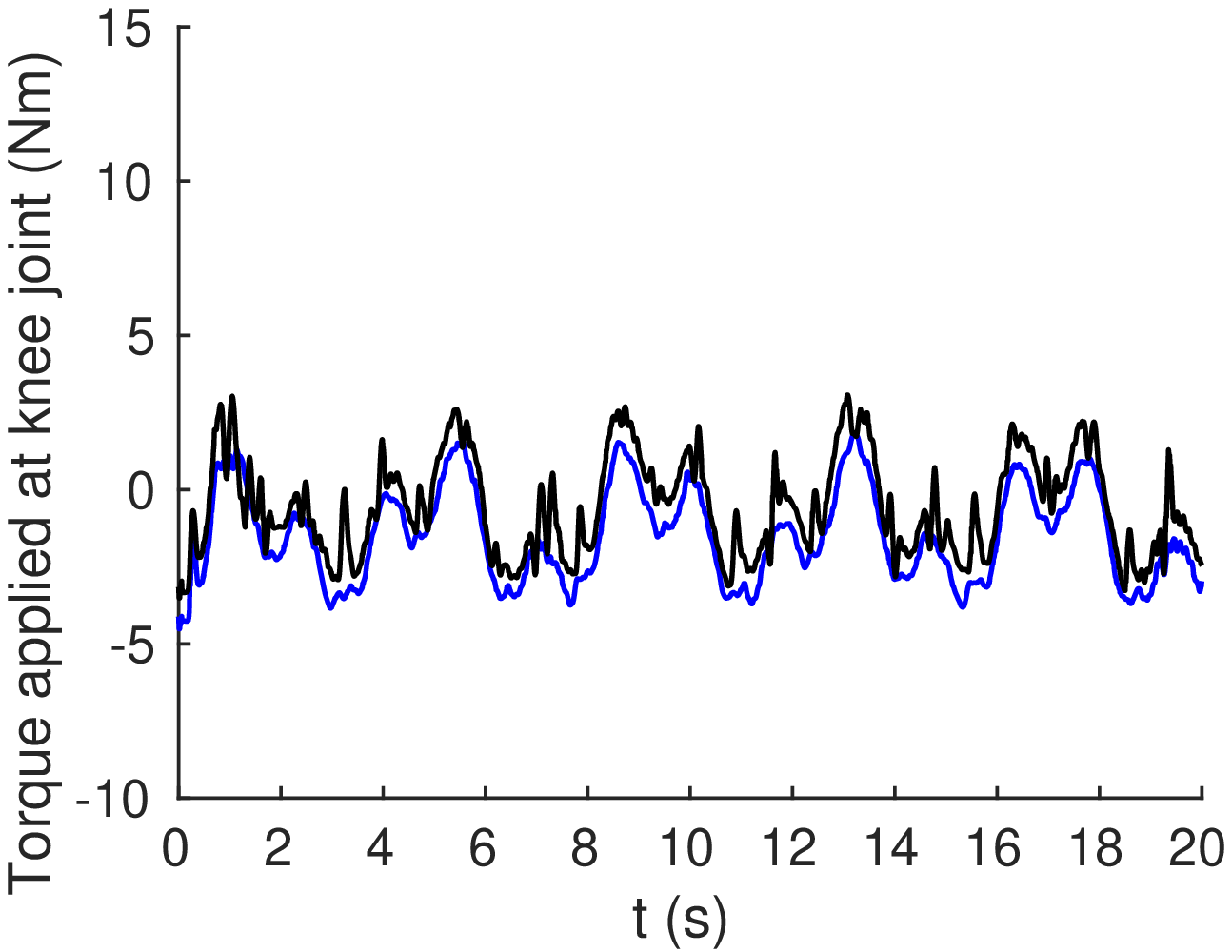}
    \label{fig_kneeSinusTorques}}
    \caption{Hip and knee joint trajectories and torques, resulting from Experiment 2. Refer to fig.~\ref{fig:ExperimentSetPoint} for legend: green lines are used for the reference joint trajectories, blue lines denote results of classical control law and black lines results of proposed control law.}
    \label{fig_Sinus}
\end{figure}

The evolution of the joint positions and torques are shown in Fig. \ref{fig_Sinus}. Results are very similar between both control laws tested. However, it can be observed that with classical control law~\eqref{eq_GravityCompensation}, the knee joint limit is exceeded at times $7 s$, $11.5 s$ and $19 s$, while with the proposed control law~\eqref{eq_torqueBounded} the joint trajectories are kept within joint limits.

\subsection{Experiment 3 - robustness versus external forces}

As another experiment, a constant reference joint position $q_d = [80, -60] deg$ was reached from a given initial position $q(0) = [-14, -60] deg$ and initial velocity $\dot{q}(0) = [0, 0]$. An external force was then applied on the foot of the robot with increasing strength, by physical interaction with a human. In the given reference position, the leg was vertical and the foot horizontal, similar to the position in Fig.~\ref{fig:iCub}. Thus, pushing on the sole of the foot was equivalent to applying an upward vertical force on the leg, affecting the hip joint.

The proportional and derivative gain matrices $K_p$ and $K_d$ were chosen as diagonal matrices with stiffness values of $[68, 17] N/m$ associated to hip and knee joints respectively, and damping values of $[0, 0]$.

\begin{figure}[!b]
    \centering 
    \subfloat{\includegraphics[width=1.6in, trim={0 0.1cm 0 0.1cm},clip]{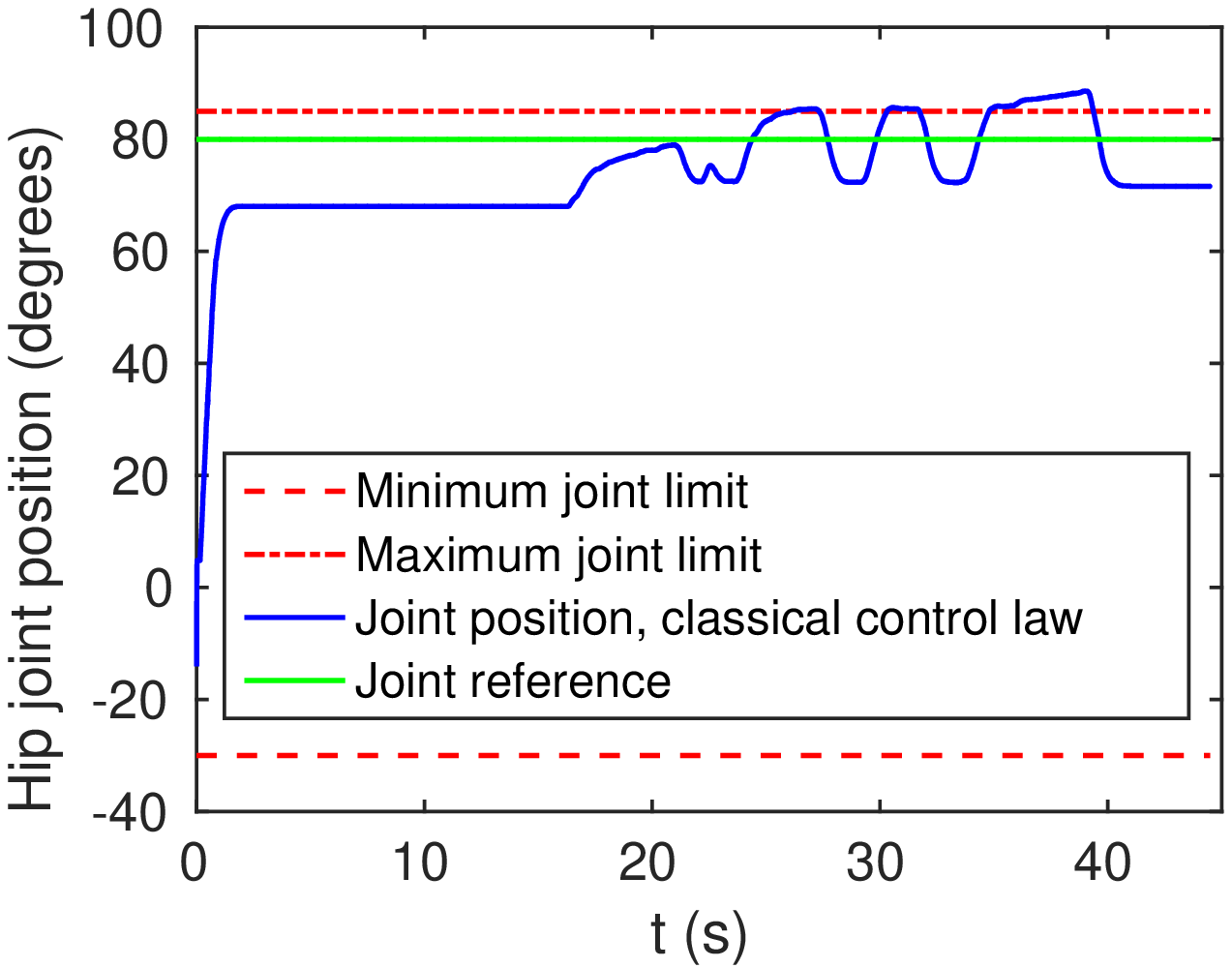}
    \label{fig_hipExtForcePositions_noVarChange}}
    ~
    \subfloat{\includegraphics[width=1.6in, trim={0 0.1cm 0 0.1cm},clip]{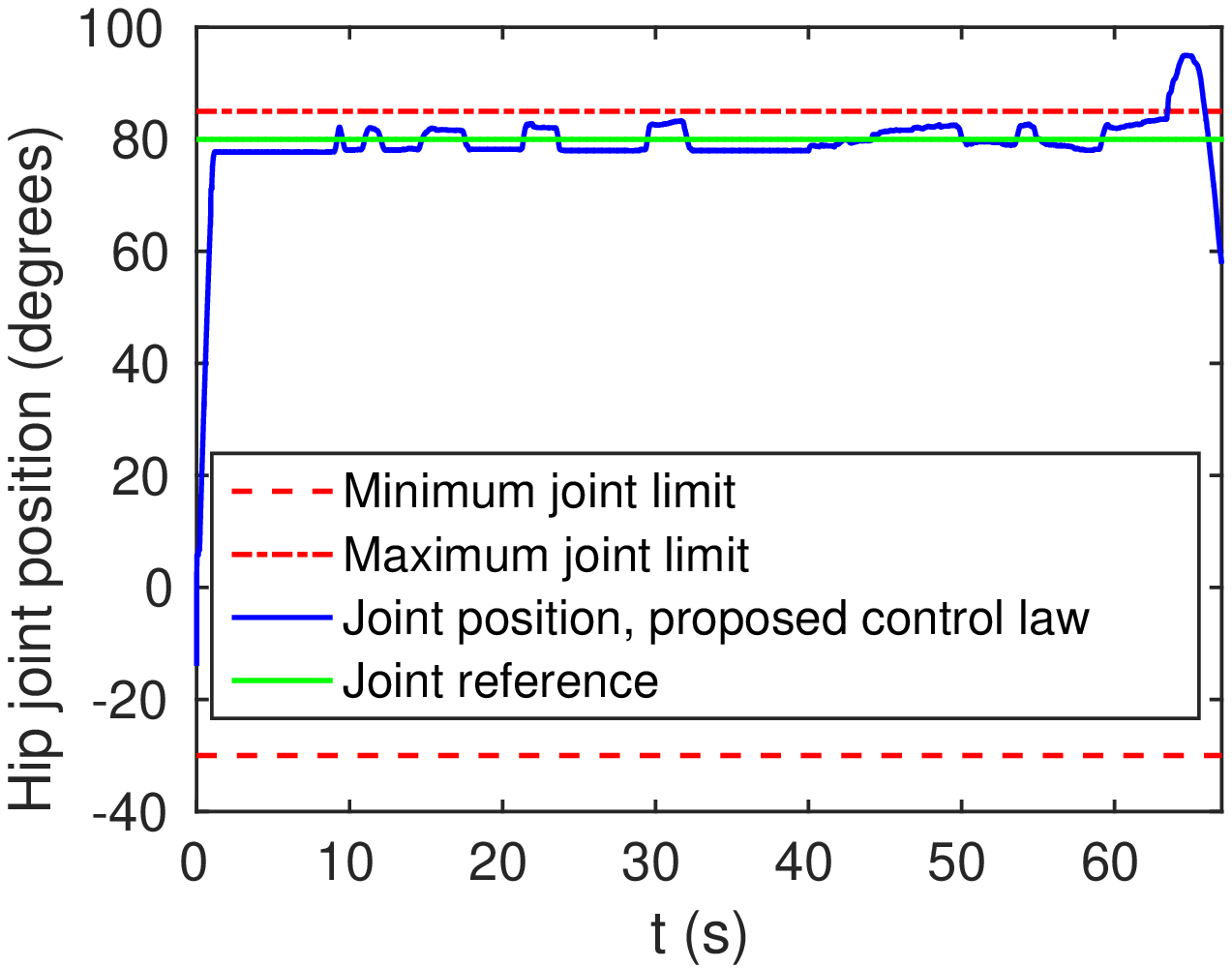}
    \label{fig_hipExtForcePositions_varChange}}
    \hfil
    \subfloat{\includegraphics[width=1.6in, trim={0 0.1cm 0 0.45cm},clip]{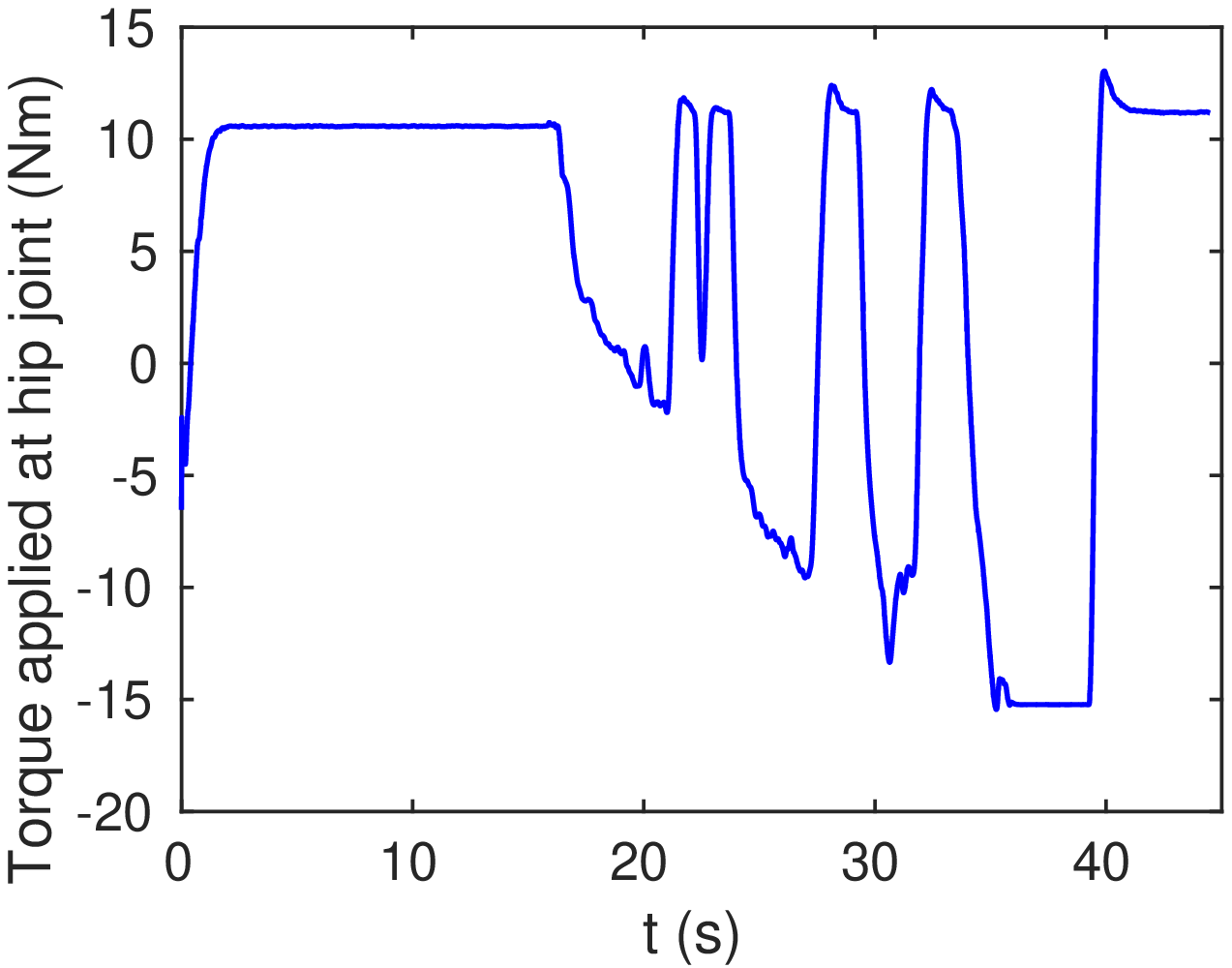}
    \label{fig_hipExtForceForce_noVarChange}}
    ~
    \subfloat{\includegraphics[width=1.6in, trim={0 0.1cm 0 0.45cm},clip]{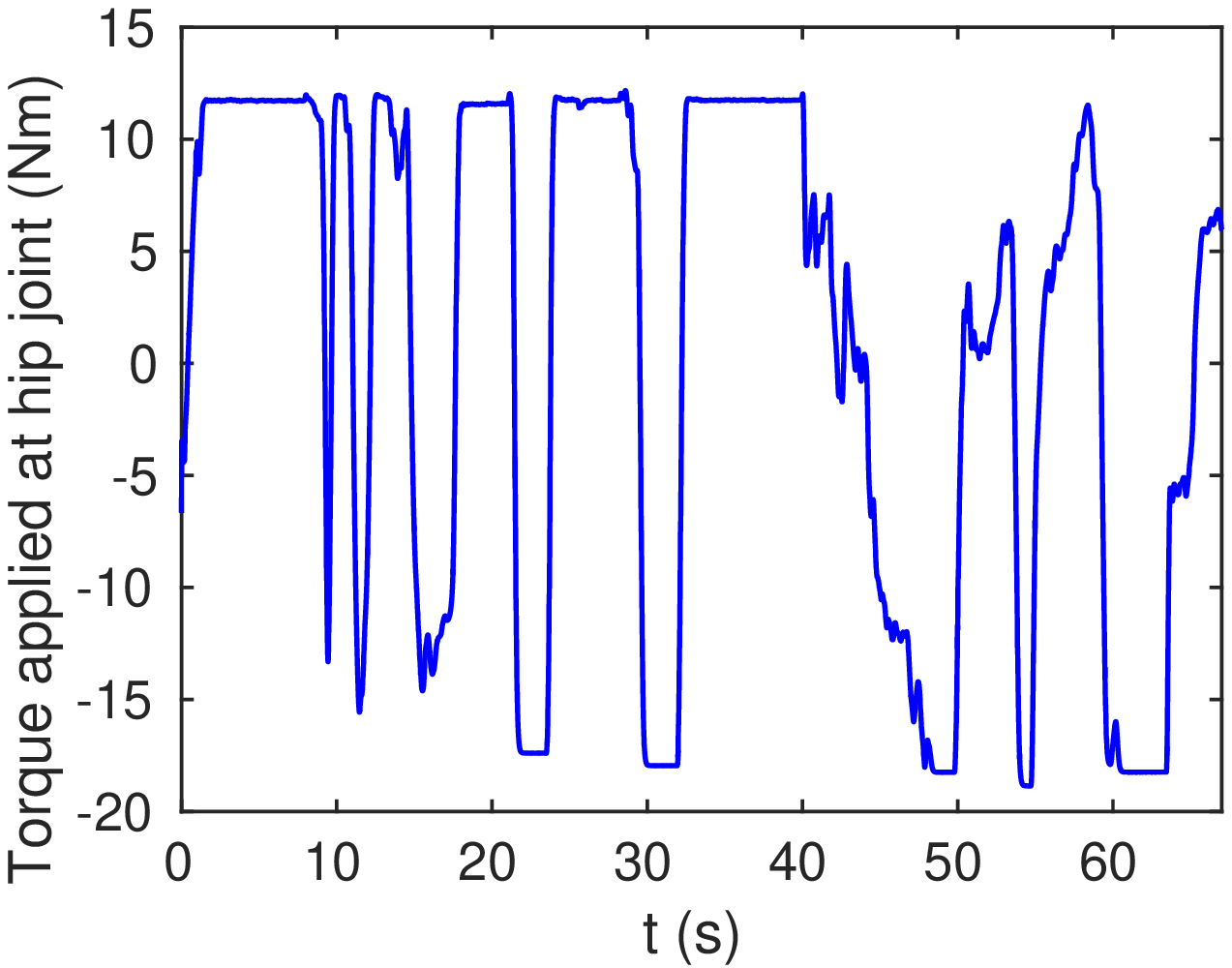}
    \label{fig_hipExtForceForce_varChange}}
    \hfil 
    \subfloat{\includegraphics[width=1.6in, trim={0 0.1cm 0 0.6cm},clip]{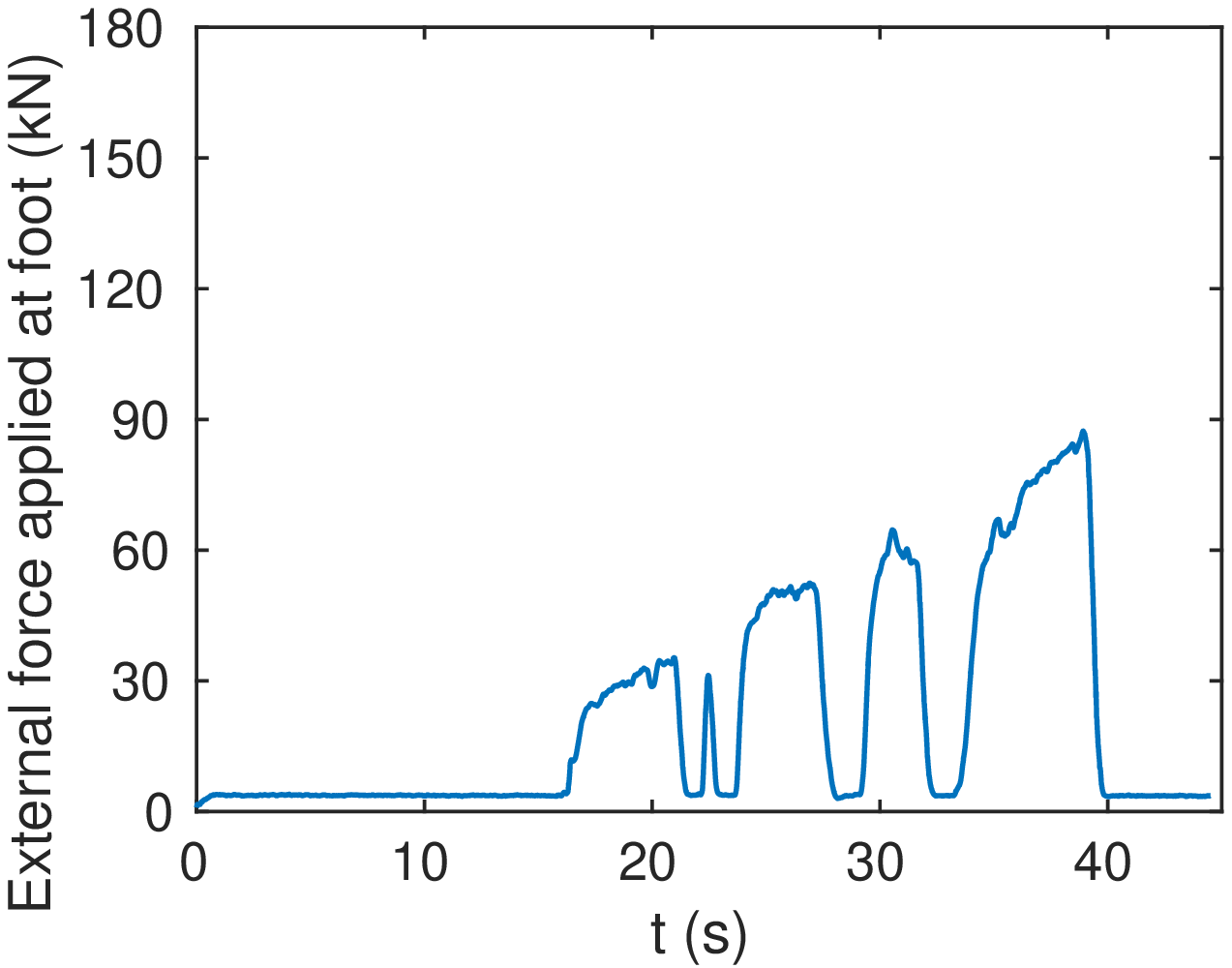}
    \label{fig_hipExtForceTorque_noVarChange}}
    ~
    \subfloat{\includegraphics[width=1.6in, trim={0 0.1cm 0 0.6cm},clip]{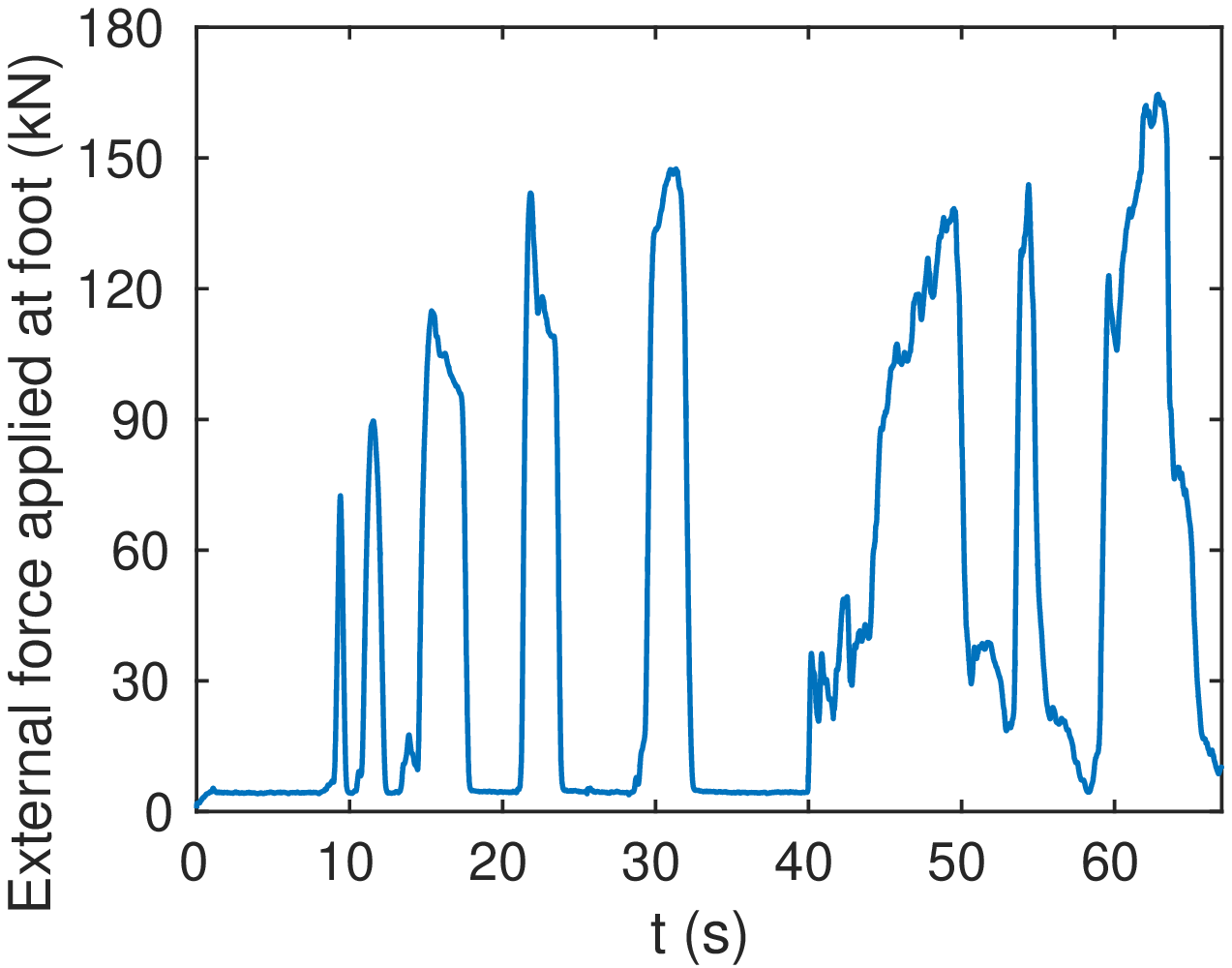}
    \label{fig_hipExtForceTorque_varChange}}
    \caption{Hip joint trajectories and torques, as well as applied external forces, for Experiment 3. On the left: results obtained with classical control law. On the right: results obtained with proposed control law.}
    \label{fig_ExtForce}
\end{figure}

Fig.~\ref{fig_ExtForce} shows the evolution of the hip joint position and control torque, as well as the vertical force applied on the foot, during the experiment. Using the classical control law~\eqref{eq_GravityCompensation}, and applying a force of $50 N$ was sufficient to move the hip position over its limit. On the other hand, a larger force of $160 N$ was applied in order to overpass the hip joint limit, when using the proposed control law~\eqref{eq_torqueBounded}.

\section{CONCLUSION AND PERSPECTIVES}

This paper presents a new joint limit avoiding controller for torque-controlled manipulators. It allows the asymptotic stabilization and convergence of a joint reference trajectory, while ensuring that the joint positions remain within their feasible range. Stability and convergence of the tracking error were shown by analysis based on Lyapunov theory. 

The approach was verified experimentally by controlling two degrees-of-freedom of the torque-controlled humanoid robot iCub. In comparison with existing passivity-based control methods, the proposed approach shows higher robustness to external perturbations, without loss of compliance: the controlled robot could resist, without overpassing joint limits, to the application of external forces 3 times larger than when controlled with a classical passivity-based control law.

The approach, in essence consisting of a change of variables, is general enough to be applied to any torque-controlled robot subject to joint limits. As future work, it is planned to extend and implement this approach into whole-body balancing control~\cite{Nava2016} of the iCub humanoid robot.

\section*{APPENDIX}


Consider the following candidate Lyapunov function:
$V:=~\frac{1}{2} \dot{\tilde{\xi}}^T M_\xi \dot{\tilde{\xi}} +  \frac{1}{2} \tilde{\xi}^T K_P \tilde{\xi}$.


Observe that $V = 0 \iff (\dot{\tilde{\xi}}, \tilde{\xi}) = (0_n, 0_n)$. Note that $K_P$ being a positive definite matrix, and in view of Property~\ref{hp:mxiPositiveDefinite}, then 
$V(\tilde{\xi},\dot{\tilde{\xi}},t) > 0 \quad \forall (\tilde{\xi},\dot{\tilde{\xi}}) - \{0\}$.
Now, recall that $M_\xi$ tends to zero when $\tilde{\xi}$ tends to infinity. Despite this fact, one shows that the candidate Lyapunov function is radially unbounded, i.e. 
$|(\tilde{\xi},\dot{\tilde{\xi}}) |\rightarrow \infty \Rightarrow V \rightarrow \infty$,
a sufficient condition for obtaining global stability results associated with a candidate Lyapunov function~\cite[p. 152]{Khalil}. This is the main point of the proof, where it differs consistently from the proof of the passivity-based controller~\eqref{eq_GravityCompensation}.

Then, in view of Property~\ref{hp:mM2CxiSkew}, the time derivative of $V$ along the closed loop system~\eqref{eq_SystemModelXi}-\eqref{eq_controlTauXi} is given by
    $\dot{V}~=~-~\dot{\tilde{\xi}}^T K_D \dot{\tilde{\xi}} \leq 0$,
which implies the stability of the equilibrium point $(\tilde{\xi},\dot{\tilde{\xi}})=(0,0)$, and boundedness of the system trajectories $(\tilde{\xi},\dot{\tilde{\xi}})(t)$ for any initial condition.

Now, observe that the closed-loop system~\eqref{eq_SystemModelXi}-\eqref{eq_controlTauXi} is time varying, and this implies that LaSalle's lemma cannot be applied to determine that $\dot{V}$ tends to zero. To show this, we have to apply Barbalat's lemma, and thus we have to show that $\ddot{V}$ is bounded. By using the fact that the trajectories of the system $(\tilde{\xi},\dot{\tilde{\xi}})(t)$ are bounded, one shows that $\ddot{V}$ is bounded. Then, $\dot{V}$ tends to zero, and this implies that $\dot{\tilde{\xi}}$ tends to zero. To show that also $\tilde{\xi}$ tends to zero, we have to show first that $\ddot{\tilde{\xi}}$ tends to zero. This latter fact can be shown by using again Barbalat's lemma, i.e. one shows that $\dddot{\tilde{\xi}}$ is bounded using the fact that the system trajectories are bounded. Then, one has $\dot{\tilde{\xi}} \rightarrow 0$ and  $\ddot{\tilde{\xi}} \rightarrow 0$. By using these facts in the closed loop dynamics~\eqref{eq_SystemModelXi}-\eqref{eq_controlTauXi}, one has that $\tilde{\xi}$ tends to zero.

\section*{ACKNOWLEDGMENT}

This work was supported the FP7 European project CoDyCo under Grant 600716 ICT 2011.2.1 Cognitive Systems and Robotics and by the Innovative Training Network SECURE, funded by the Horizon 2020 Marie Sklodowska Curie Actions (MCSA) of the European Commission (H2020-MSCA-ITN-2014- 642667)







\end{document}